\documentclass[runningheads]{llncs}
\usepackage{graphicx}
\usepackage{comment}
\usepackage{amsmath,amssymb} 
\usepackage{color}
\usepackage[breaklinks=true,bookmarks=false]{hyperref}
\usepackage{epsfig}
\usepackage{graphicx}
\usepackage{amsmath}
\usepackage{amssymb}
\usepackage{booktabs}
\usepackage{adjustbox}
\usepackage{enumitem}
\usepackage{multirow}
\usepackage{marvosym}
\usepackage{caption}
\usepackage{algorithm,algorithmic,amsmath}
\usepackage{tabularx}
\usepackage{colortbl}
\usepackage{pifont}
\definecolor{Gray}{gray}{0.9}

\begin{document}
\pagestyle{headings}
\mainmatter
\def\ECCVSubNumber{781}  

\title{Structured Landmark Detection via Topology-Adapting Deep Graph Learning} 


\titlerunning{Landmark Detection via Topology-Adapting Deep Graph Learning}
%
\author{Weijian Li\inst{1,2} \and
Yuhang Lu\inst{1,3} \and
Kang Zheng\inst{1} \and
Haofu Liao\inst{2} \and
Chihung Lin\inst{4} \and
Jiebo Luo\inst{2} \and
Chi-Tung Cheng\inst{4} \and
Jing Xiao\inst{5} \and
Le Lu\inst{1} \and
Chang-Fu Kuo\inst{4} \and
Shun Miao\inst{1}}

\authorrunning{Li et al.}
%
\institute{PAII. Inc., Bethesda, MD, USA \\
\and
Department of Computer Science, University of Rochester, Rochester, NY, USA \\
\and
Department of Computer Science and Engineering, University of South Carolina, Columbia, SC, USA \\ 
\and
Chang Gung Memorial Hospital, Linkou, Taiwan, ROC \\
\and
Ping An Technology, Shenzhen, China \\ 
}
\maketitle
\begin{abstract}

Image landmark detection aims to automatically identify the locations of predefined fiducial points. Despite recent success in this field, higher-ordered structural modeling to capture implicit or explicit relationships among anatomical landmarks has not been adequately exploited. In this work, we present a new topology-adapting deep graph learning approach for accurate anatomical facial and medical (e.g., hand, pelvis) landmark detection. The proposed method constructs graph signals leveraging both local image features and global shape features. The adaptive graph topology naturally explores and lands on task-specific structures which are learned end-to-end with two Graph Convolutional Networks (GCNs). Extensive experiments are conducted on three public facial image datasets (\textit{WFLW, 300W, and COFW-68}) as well as three real-world X-ray medical datasets (\textit{Cephalometric (public), Hand and Pelvis}). Quantitative results comparing with the previous state-of-the-art approaches across all studied datasets indicating the superior performance in both robustness and accuracy. Qualitative visualizations of the learned graph topologies demonstrate a physically plausible connectivity laying behind the landmarks.

\end{abstract}

\section{Introduction}
Image landmark detection has been a fundamental step for many high-level computer vision tasks to extract and distill important visual contents, such as image registration \cite{han2015robust}, pose estimation \cite{bulat2017binarized}, identity recognition \cite{zhu2013deep} and image super-resolution \cite{bulat2018super}. Robust and accurate landmark localization becomes a vital component determining the success of the downstream tasks.

Recently, heatmap regression based methods~\cite{wu2018look,zhu2019robust,sun2019deep,valle2018deeply} have achieved encouraging performance on landmark detection. They model landmark locations as heatmaps and train deep neural networks to regress the heatmaps. Despite popularity and success, they usually suffer from a major drawback of lacking a global representation for the structure/shape, which provides high-level and reliable cues in individual anatomical landmark localization. As a result, heatmap-based methods could make substantial errors when being exposed to large appearance variations such as occlusions.

In contrast, coordinate regression based methods~\cite{lv2017deep,zhang2015learning,yu2016deep,trigeorgis2016mnemonic} have an innate potential to incorporate structural knowledge since the landmark coordinates are directly expressed. Most existing methods initialize landmark coordinates using mean or canonical shapes, which indirectly inject weak structural knowledge \cite{trigeorgis2016mnemonic}. While the exploitation of the structural knowledge in existing methods has still been insufficient as well as further exploitation of the structural knowledge considering the underlying relationships between the landmarks. Effective means for information exchange among landmarks to facilitate landmark detection are also important but have yet to be explored. Due to these limitations, the performance of the latest coordinate-based methods \cite{wu2017leveraging} falls behind the heatmap-based ones \cite{wang2019adaptive}.

In this work, we introduce a new topology-adapting deep graph learning approach for landmark detection, termed \textit{Deep Adaptive Graph (DAG)}. We model the landmarks as a graph and employ global-to-local cascaded Graph Convolutional Networks (GCNs) to move the landmarks towards the targets in multiple steps. Graph signals of the landmarks are built by combining local image features and graph shape features. Two GCNs operate in a cascaded manner, with the first GCN estimating a global transformation of the landmarks and the second GCN estimating local offsets to further adjust the landmark coordinates. The graph topology, represented by the connectivity weights between landmarks, are learned during the training phase. 

By modeling landmarks as a graph and processing it with GCNs, our method is able to effectively exploit the structural knowledge and allow rich information exchange among landmarks for accurate coordinate estimation. The graph topology learned for landmark detection task is capable of revealing reasonable landmark relationships for the given task. It also reduces the need for manually defining landmark relations (or grouping), making our method to be easily adopted for different tasks. By incorporating shape features into graph signal in addition to the local image feature, our model can learn and exploit the landmark shape prior to achieve high robustness against large appearance variations (e.g., occlusions). 
In summary, our main contributions are four-fold: 
\begin{enumerate}
  \item By representing the landmarks as a graph and detecting them using GCNs, our method effectively exploits the structural knowledge for landmark coordinate regression, closes the performance gap between coordinate- and heatmap-based landmark detection methods.
  \item Our method automatically reveals physically meaningful relationships among landmarks, leading to a task-agnostic solution for exploiting structural knowledge via step-wise graph transformations.
  \item Our model combines both visual contextual information and spatial positional information into the graph signal, allowing structural shape prior to be learned and exploited.
  \item Comprehensive quantitative evaluations and qualitative visualizations on six datasets across both facial and medical image domains demonstrate the consistent state-of-the-art performance and general applicability of our method.
\end{enumerate}


\section{Related Work}
A large number of studies have been reported in this domain including the classic Active Shape Models \cite{milborrow2008locating,cootes1995active,cootes1992active}, Active Appearance Models \cite{cootes2001active,sauer2011accurate,liu2007generic}, Constraind Local Models \cite{cristinacce2006feature,asthana2013robust,saragih2009face,lindner2014robust}, and more recently the deep learning based models which can be further categorized into heatmap or regression based models.

\textbf{Heatmap Based Landmark Detection:}  These methods \cite{wei2016convolutional,newell2016stacked,tang2018quantized,sun2019deep,payer2016regressing,chen2019cephalometric} generate localized predictions of likelihood heatmaps for each landmark and achieve encouraging performances. A preliminary work by Wei \textit{et al.} \cite{wei2016convolutional} introduce a Convolutional Pose Machine (CPM) which models the long-range dependency with a multistage network. Newell \textit{et al.} \cite{newell2016stacked} propose a Stacked Hourglass model leveraging the repeated bottom-up and top-down structure and intermediate supervision. Tang \textit{et al.} \cite{tang2018quantized} investigate a stacked U-Net structure with dense connections. Lately, Sun \textit{et al.} \cite{sun2019deep} present a deep model named High-Resolution Network (HRNet18) which extracts feature maps in a joint deep and high resolution manner via conducting multi-scale fusions across multiple branches under different resolutions. Based on these models, other methods also integrate additional supervision cues such as the object structure constraints \cite{wu2019facial,zoulearning}, the variety of image, and object styles \cite{dong2018style,qian2019aggregation} to solve specific tasks.

\textbf{Coordinate Based Landmark Detection:} Another common approach directly locates landmark coordinates from input images~\cite{toshev2014deeppose,sun2013deep,trigeorgis2016mnemonic,lv2017deep,zhu2015face,liu2016fashion,su2019efficient}. Most of these methods consist of multiple steps to progressively update predictions based on visual signals, widely known as Cascaded-Regression. Toshev \textit{et al.} \cite{toshev2014deeppose} and Sun \textit{et al.} \cite{sun2013deep} adopt cascaded Convolutional Neural Networks (CNNs) to predict landmark coordinates. Trigeorgis \textit{et al.} \cite{trigeorgis2016mnemonic} model the cascaded regression process using a Recurrent Neural Network (RNN) based deep structure. Lv \textit{et al.} \cite{lv2017deep} propose a two-stage regression model with global and local reinitializations. From different perspectives, Zhu \textit{et al.} \cite{zhu2015face} investigate the methods of optimal initialization by searching the object shape space; Valle \textit{et al.} \cite{valle2018deeply} present a combined model with a tree structured regressor to infer landmark locations based on heatmap prediction results; Wu \textit{et al.} \cite{wu2017leveraging} leverage uniqueness and discriminative characteristics across datasets to assist landmark detection.

\textbf{Landmark Detection with Graphs:} The structure of landmarks can be naturally modeled as a graph considering the landmark locations and landmark to landmark relationships \cite{zhou2013exemplar,yu2019layout,saragih2009face,yu2015face,zhu2015face}. Zhou \textit{et al.} \cite{zhou2013exemplar} propose a Graph-Matching method which obtains landmark locations by selecting the set of landmark candidates that would best fit the shape constraints learned from the examplars. Yu \textit{et al.} \cite{yu2015face} describe a two-stage deformable shape model to first extract a coarse optimum by maximizing a local alignment likelihood in the region of interest then refine the results by maximizing an energy function under shape constraints. Later, Yu \textit{et al.} \cite{yu2019layout} present a hierarchical model to extract semantic features by constructing intermediate graphs from bottom-up node clustering and top-down graph deconvolution operations, leveraging the graph layout information. Zou \textit{et al.} \cite{zoulearning} introduce a landmark structure construction method with covering set algorithm. While their method is based on heatmap detection results, we would like to directly regress landmark locations from raw input image to avoid potential errors incurred from heatmap detections.

Recently, Ling \textit{et al.} \cite{ling2019fast} propose a fast object annotation framework, where contour vertices are regressed using GCN to perform segmentation, indicating the benefit of position prediction with iterative message exchanges. In their task, each point is considered with the same semantics towards coarse anonymous matching which is not appropriate for precise targeted localization tasks like landmark detection. Adaptively learning graph connectivities instead of employing a fixed graph structure based on prior knowledge should be explored to improve the model's generalizability to different tasks.

\section{Method}

\begin{figure*}[t!]
	\centering
	\includegraphics[width=\textwidth]{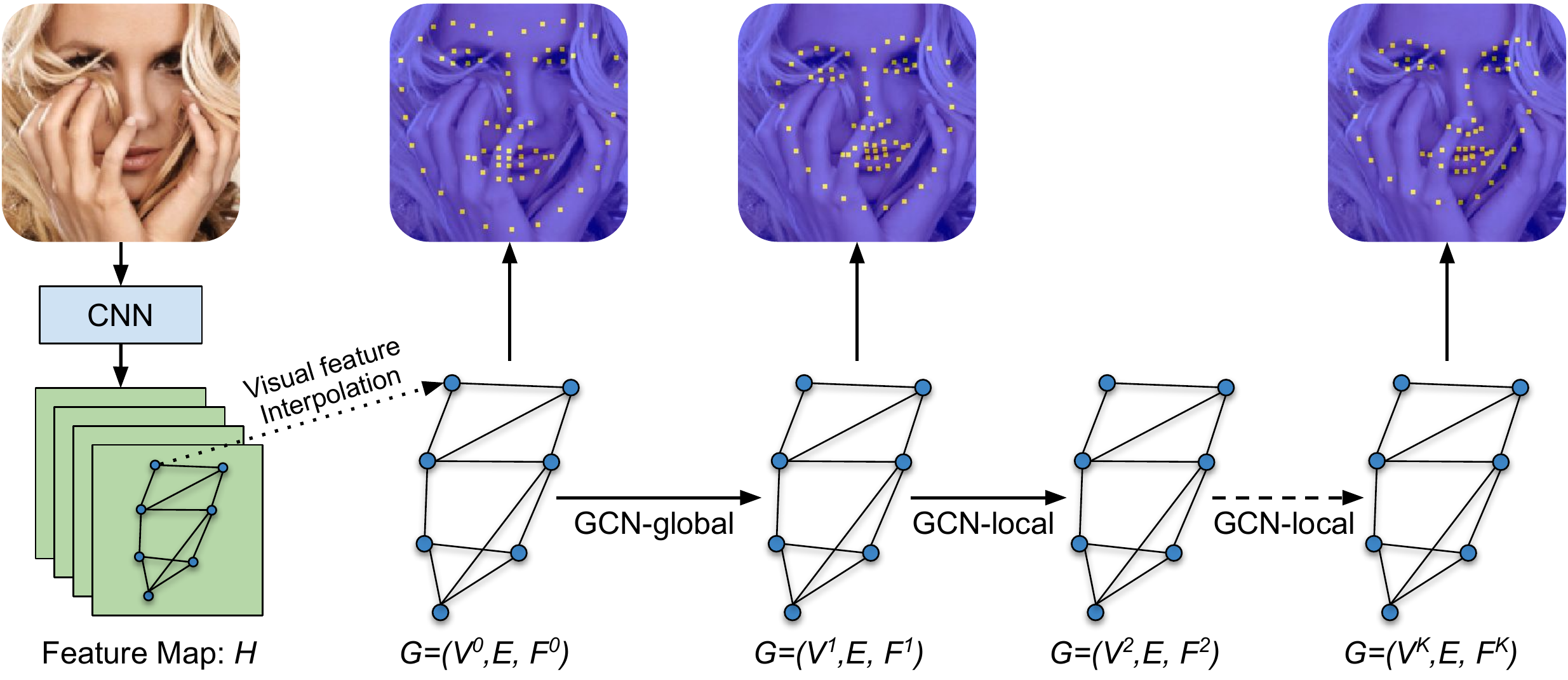}
	\caption{Overview of the proposed Deep Adaptive Graph (DAG). Initial graph is initialized with the mean value computed from training data. We first deform the landmark graph through a perspective transformation predicted by GCN-global and then precisely shift the graph by GCN-local through iterations. The visual features and shape features are re-interpolated from feature map and re-calculated after each GCN module, respectively.}
	\label{fig:framework}
\end{figure*}

Our method adopts the cascaded-regression framework, where given the input image and initial landmarks (from the mean shape), the predicted landmark coordinates are updated in multiple steps. Yet differently, we feature the cascaded-regression framework with a graph representation of the landmarks, denoted by $G=(V, E, F)$, where $V=\{\mathbf{v}_i\}$ denotes the landmarks, $E=\{e_{ij}\}$ denotes the learned connectivity between landmarks and $F=\{\mathbf{f}_i\}$ denotes graph signals capturing appearance and shape information. The graph is processed by cascaded GCNs to progressively update landmark coordinates. An overview of our method is shown in Figure~\ref{fig:framework}. Details of the cascaded GCNs, graph signal and learned connectivity are presented in Section \ref{subsec:gcn}, Section \ref{subsec:graphsignal} and Section \ref{subsec:graphadj}, respectively. The training scheme of our method can be found in Section \ref{subsec:training}.

\subsection{Cascaded GCNs}
\label{subsec:gcn}

Given a graph representation of landmarks $G=(V, E, F)$, two-stage cascaded GCN modules are employed to progressively update the landmark coordinates. The first stage, \textit{GCN-global}, estimates a global transformation to coarsely move the landmarks to the targets. The second stage, \textit{GCN-local}, estimates local landmark coordinate offsets to iteratively move the landmarks toward the targets. Both modules employ the same GCN architecture (weights not shared) and the same learnable graph connectivity. 

\textbf{Graph Convolution:} 
Given a graph connectivity $E$ and a graph feature $F$, the $k$-th graph convolution operation updates the $i$-th node feature $\mathbf{f}^j_k$ by aggregating all node features weighted by the connectivity:
\begin{equation}
	\mathbf{f}^i_{k+1} = \mathbf{W}_1\mathbf{f}^i_k + \sum_{j}e_{ij} \mathbf{W}_2\mathbf{f}^j_k
	\label{eqn:graph_conv}
\end{equation}
where $\mathbf{W}_1$ and $\mathbf{W}_2$ are learnable weight matrices. The graph convolutions can be seen as the mechanism of information collection among the neighborhoods. The connectivity $E$ serves as pathways for information flow from one landmark to another. 

\textbf{Global Transformation GCN:} 
Previous work \cite{jaderberg2015spatial,lv2017deep} learn an affine transformation with a deep neural network by predicting a two by three affine transformation matrix which deforms the image to the satisfied posture. Inspired by this work, we employ a GCN on the initial landmarks to coarsely move them to the targets. Considering our graph is more flexible that does not have to maintain the parallelism and respective ratios among the edges, we model the global transformation using a perspective transformation~\cite{detone2016deep}. A perspective transformation can be parameterized by 9 scalars $M=[a,b,c,d,e,f,g,h,i]^T \in\mathbb{R}^{9 \times 1}$ with the operation written as:
\begin{equation}
 \left[ \begin{array}{c}
x' \\
y' \\
1 \end{array} \right]
\cong \left[ \begin{array}{c}
rx' \\
ry' \\
r \end{array} \right] 
= \left( \begin{array}{ccc}
a & b & c \\
d & e & f \\
g & h & i \end{array} \right)
\left[ \begin{array}{c}
x \\
y \\
1 \end{array} \right]
\end{equation}

Given a target image, we initialize landmark locations $V^0$ using the mean shape of landmarks in the training set, and placed it at the center of the image. The graph is processed by the GCN-global to estimate a perspective transformation to bring the initial structure closer to the target. 

Specifically, a graph isomorphism network (GIN)~\cite{xu2018powerful} is employed to process the graph features $\{ \mathbf{f}^i_k \}$ produced by the GCN to output a 9-dimensional vector representing the perspective transformation:
\begin{equation}
	\mathbf{f}^G = \mbox{MLP}\left ( \mbox{CONCAT}\left ( \mbox{READOUT}\left(\left \{ \mathbf{f}^i_k | i \in G \right \} \right) | k=0,1,\dots,K \right) \right ),
\end{equation}
where the READOUT operator sums the features from all the nodes in the graph $G$. The transformation matrix $M$ is obtained by transforming and reshaping $\mathbf{f}^G$ into a 3 by 3 matrix. We then apply this transformation matrix on the initial landmark node coordinates to obtain the aligned landmark coordinates:
\begin{equation}
    V^1 = \{\mathbf{v}^1_i\} = \{ \mathbf{M}\mathbf{v}^0_i\}
\end{equation}

\textbf{Local Refinement GCN:}
Given the transformed landmarks, we employ GCN-local to further shift the graph in a cascaded manner.
GCN-local employs the same architecture as GCN-global, with a difference that the last layer produces a 2-dimensional vector for each landmark, representing the coordinate offset of the landmark. The updated landmark coordinates can be written as:
\begin{equation}
	\mathbf{v}^{t+1}_i = \mathbf{v}^{t}_i + \Delta \mathbf{v}^{t}_i,
	\label{eq:step-shift}
\end{equation}
where $\Delta \mathbf{v}_i^{t}=(\Delta x_i^{t}, \Delta y_i^{t})$ is the output of the GCN-local at the $t$-th step. In all our experiments, we perform $T=3$ iterations of the GCN-local. Note that the graph signal is re-calculated after each GCN-local iteration. 


\subsection{Graph signal with appearance and shape information}
\label{subsec:graphsignal}

We formulate a graph signal $F$ as a set of node features $\mathbf{f}_i$, each associated with a landmark $\mathbf{v}_i$. The graph signal contains a \textit{visual feature} to encode local image appearance and a \textit{shape feature} to encode the global landmark shape. 

\textbf{Visual Feature:} Specifically, given a feature map $H$ with $D$ channels produced by a backbone CNN, visual features, denoted by $\mathbf{p}_i \in R^D$, are extracted by interpolating $H$ at the landmark coordinates $\mathbf{v}_i$. The interpolation is performed via a differentiable bi-linear interpolation \cite{jaderberg2015spatial}. In this way, visual feature of each landmark is collected from the feature map, encoding the appearance of its neighborhood.

\textbf{Shape Feature:} While the visual feature encodes the appearance in a neighborhood of the landmark, it does not explicitly encode the global shape of the landmarks. To incorporate this structural information into the graph signal, for each landmark, we compute its displacement vectors to all other landmarks, denoted as $\mathbf{q}_i = \{\mathbf{v}_j - \mathbf{v}_i\}_{j \neq i} \in R^{2 \times (N-1)}$, where $N$ is the number of landmarks. Such shape feature allows structural information of the landmarks to be exploited to facilitate landmark detection. For example, when the mouth of a face is occluded, the coordinates of the mouth landmarks can be inferred from the eyes and nose. Wrong landmark detection results that violate the shape prior can also be avoided when the shape is explicitly captured in the graph signal. 

The graph signal $F$ is then constructed for each landmark by concatenating the visual feature $\mathbf{p}_i$ and the shape feature $\mathbf{q}_i$ (flattened), resulting in a feature vector $\mathbf{f}_i \in R^{D + 2(N-1)}$.

\subsection{Landmark graph with learnable connectivity}
\label{subsec:graphadj}

The graph connectivity determines the relationship between each pair of landmarks in the graph and serves as the information exchange channel in GCN. In most existing applications of GCN~\cite{qi2019attentive,ling2019fast,zhao2019semantic,velivckovic2017graph,wu2019session}, the graph connectivity is given based on the prior knowledge of the task. In our landmark detection application, it is non-trivial to manually define the optimal underlying graph connectivity for the learning task. 
Therefore, relying on hand-crafted graph connectivity would introduce a subjective element into the model, which could lead to sub-optimal performance. To address this limitation, we learn task-specific graph connectivities during the training phase in an end-to-end manner. The connectivity weight $e_{ij}$ behaves as information propagation gate in graph convolutions (Eqn. \ref{eqn:graph_conv}). We treat the connectivity $\{e_{ij}\}$, represented as an adjacency matrix, as a learnable parameter that is trained with the network during the training phase. In this way, the task-specific optimal graph connectivity is obtained by optimizing the performance of the target landmark detection task, allowing our method to be applied to different landmark detection tasks without manual intervention.

Graph connectivity learning has been studied before by the research community. One notable example is Graph Attention Networks~\cite{velivckovic2017graph}, which employs a self-attention mechanism to adaptively generate connectivity weights during the model inference. We conjugate that in structured landmark detection problems, the underlying relationship between the landmarks remains the same for a given task, instead of varying across individual images. Therefore, we share the same connectivity across images on the same task, and directly optimize the connectivity weights during the training phase.

\subsection{Training}
\label{subsec:training}

\textbf{GCN-global:} Since the perspective transformation estimated by GCN-global has limited degree of freedom, directly penalizing the distance between the predicted and the ground truth landmarks will lead to unstable optimization behavior. As the goal of GCN-global is to coarsely locate the landmarks, we propose to use a margin loss on the $L_1$ distance, written as:
\begin{equation}
	\mathcal{L}_{global}= \left[\left(\frac{1}{N}\sum_{i\in N}\sum_{x,y}|\mathbf{v}_i^1-\mathbf{v}_i|\right)-m\right]_+
\end{equation}
where $[u]_+ := max(0, u)$. $\mathbf{v}_i^1=(x_i^1, y_i^1)$ and $\mathbf{v}_i=(x_i, y_i)$ denote the predicted and ground truth landmark coordinates for the $i$-th landmark. $m$ is a hyper-parameter representing a margin which controls how well we want the alignment to be. Following this procedure, we aim to obtain a high robustness of the coarse landmark detection, while forgive small errors.

\textbf{GCN-local:} To learn a precise localization, we directly employ L1 loss on all predicted landmark coordinates after the GCN-local, written as:
\begin{equation}
	\mathcal{L}_{local}= \frac{1}{N}\sum_{i\in N}\sum_{x,y}|\mathbf{v}_i^{T}-\mathbf{v}_i|
\end{equation}
where $\mathbf{v}_i^{T}$ is the $T$-th step (the last step) coordinate predictions, and $\mathbf{v}_i$ is the ground truth coordinate for the $i$-th landmark.

The overall loss to train DAG is a combination of the above two losses:
\begin{equation}
	\mathcal{L}= \lambda_1\mathcal{L}_{global} + \lambda_2\mathcal{L}_{local}
\end{equation}
where $\lambda_k$ is the weight parameter for each loss.

\section{Experiments}
\subsection{Datasets} 

We conduct evaluations on three public facial image and three medical image datasets:

\noindent\textbf{WFLW} \cite{wu2018look} dataset contains 7,500 facial images for training and 2,500 facial images for testing. The testing set is further divided into 6 subsets focusing on particular challenges in the images namely large pose set, expression set, illumination set, makeup set, occlusion set, and blur set. 98 manually labeled landmarks are provided for each image.

\noindent\textbf{300W} \cite{sagonas2013300} dataset consists of 5 facial datasets namely LFPW, AFW, HELEN, XM2VTS and IBUG. They are split into a training set with 3,148 images, and a testing set with 689 images where 554 images are from LFPW and HELEN, 135 from IBUG. Each image is labeled with 68 landmarks.

\noindent\textbf{COFW} \cite{burgos2013robust} dataset contains 1,345 facial images for training and 507 for testing, under different occlusion conditions. Each image is originally labeled with 29 landmarks and re-annotated with 68 landmarks \cite{ghiasi2015occlusion}. We follow previous studies \cite{wu2018look,qian2019aggregation} to conduct inferences on the re-annotated COFW-68 dataset to test our model's cross-dataset performance which is trained on 300W dataset.

\noindent\textbf{Cephalometric X-ray} \cite{wang2016benchmark} is a public dataset originally for a challenge in IEEE ISBI-2015. It contains 400 X-ray Cephalometric images with resolution of $1,935 \times 2,400$, 150 images are used as training set, the rest 150 images and 100 images are used as validation and test sets. Each cephalometric image contains 19 landmarks. In this paper, we only focus on the landmark detection task.

\noindent\textbf{Hand X-ray} \cite{lu2020learning} is a real-world medical dataset collected by a hospital. The X-ray images are taken with different hand poses with resolutions in $1,500s \times 2,000s$. In total, 471 images are randomly split into a training set (80\%, $N$=378) and a testing set (20\%, $N$=93). 30 landmarks are manually labeled for each image.

\noindent\textbf{Pelvic X-ray} \cite{wang2019weakly,chen2020anatomy} another real-world medical dataset collected by the same hospital. Images are taken over patient's pelvic bone with resolutions in $2,500s \times 2,000s$. The challenges in this dataset is the high structural and appearance variation, caused by bone fractures and metal prosthesis. In total, 1,000 imagesare randomly splited into a training set (80\%, $N$=800) and a testing set (20\%, $N$=200). 16 landmarks are manually labeled for each image. 

\begin{figure*}[t]
	\centering
	\includegraphics[width=\textwidth]{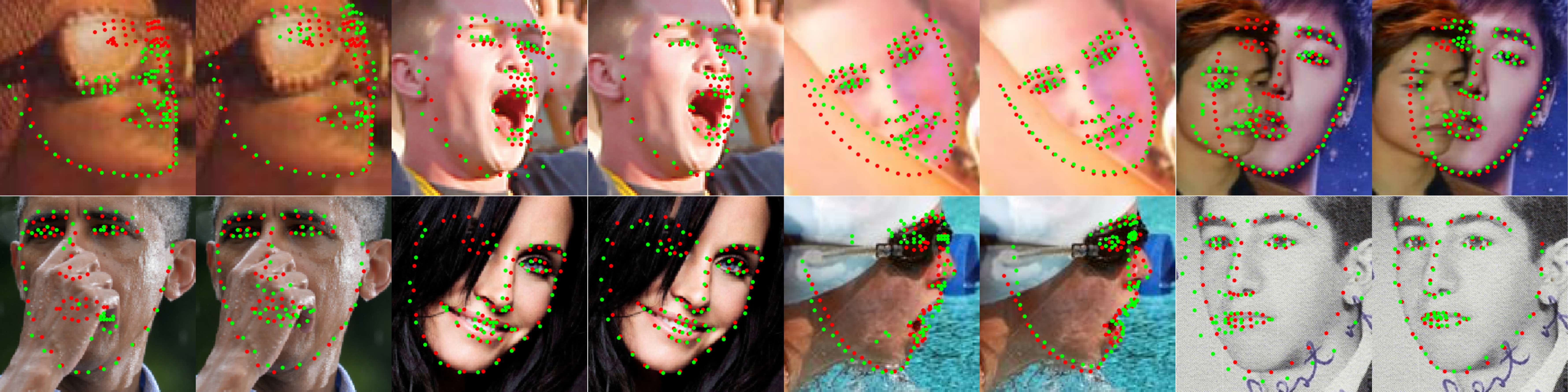}
	\includegraphics[width=\textwidth]{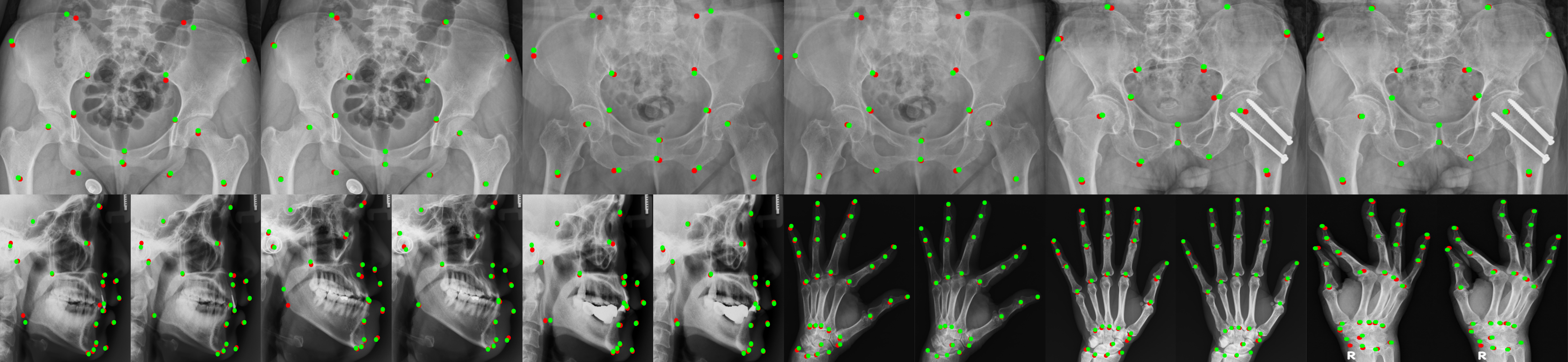}
	\caption{Visualization of landmark detection results. Pairs of reseults are displayed side by side. 
	For each pair, \textbf{Left image:} detection result from a SOTA method~\cite{sun2019deep}. \textbf{Right image:} result produced by our method. \textbf{Green dot}: predicted landmark location. \textbf{Red dot}: groundtruth landmark location.
	}
	\label{fig:face_visual}
\end{figure*}

\subsection{Experiment Settings}

\textbf{Evaluation Metrics:} We evaluate the proposed method following two sets of metrics. For the facial image datasets, we employ the widely adopted Normalized Mean Error (NME), Area Under the Curve (AUC), Failure Rate for a maximum error of 0.1 (FR@0.1) and Cumulative Errors Distribution (CED) curve (supplementary material). To compare with previous methods, we conduct both "inter-ocular" (outer-eye-corner-distance) and "inter-pupil" (eye-center-distance) normalizations on the detected landmark coordinates.

For the Cephalometric X-ray images, we follow the original evaluation protocol to compare two sets of metrics: Mean Radial Error (MRE) which computes the average of Euclidean Distances of predicted coordinates and ground truth coordinates of all the landmarks; the corresponding Successful Detection Rate (SDR) under 2mm, 2.5mm, 3mm and 4mm. For the Hand and Pelvic X-rays, we compute MRE, Hausdorff Distance (HD) and Standard Deviations (STD). Recall that Hausdorff Distance measures the maximum value of the minimum distances between two sets of points. In our case, we aim to evaluate the error upper-bound for the detected landmarks.
\begin{table*}[t]
  \centering
    \caption{Evaluation on the WFLW dataset (98 Landmarks). *: focus on loss function. \#: focus on data augmentation.}
    \resizebox{0.85\linewidth}{!}{
      \begin{tabular}{ccccccccc}
        \toprule 
            Metric & Method & Test & Pose & Expression & Illumination & Make-up & Occlusion & Blur \\
        \midrule
            \multirow{8}{*}{{Mean Error \%}}
                & CFSS \cite{zhu2015face} & 9.07 & 21.36 & 10.09 & 8.30 & 8.74 & 11.76 & 9.96 \\
                & DVLN \cite{wu2017leveraging} & 6.08 & 11.54 & 6.78 & 5.73 & 5.98 & 7.33 & 6.88 \\
                & LAB \cite{wu2018look} & 5.27 & 10.24 & 5.51 & 5.23 & 5.15 & 6.79 & 6.32 \\
                & SAN \cite{dong2018style} \# & 5.22 & 10.39 & 5.71 & 5.19 & 5.49 & 6.83 & 5.80 \\
                & WING \cite{feng2018wing} * & 5.11 & 8.75 & 5.36 & 4.93 & 5.41 & 6.37 & 5.81 \\
                & HRNet18 \cite{sun2019deep} & 4.60 & 7.94 & 4.85 & 4.55 & 4.29 & 5.44 & 5.42 \\
                & STYLE  \cite{qian2019aggregation} \# & 4.39 & 8.42 & 4.68 & 4.24 & 4.37 & 5.60 & 4.86 \\
                & AWING \cite{wang2019adaptive} * & 4.36 & 7.38 & 4.58 & 4.32 & 4.27 & 5.19 & 4.96 \\
                & \textbf{Ours} & \textbf{4.21} & \textbf{7.36} & \textbf{4.49} & \textbf{4.12} & \textcolor{blue}{\textbf{4.05}} & \textcolor{blue}{\textbf{4.98}} & \textbf{4.82} \\
        \midrule
            \multirow{8}{*}{{Failure Rate @0.1}}
                & CFSS \cite{zhu2015face} & 20.56 & 66.26 & 23.25 & 17.34 & 21.84 & 32.88 & 23.67 \\
                & DVLN \cite{wu2017leveraging} & 19.84 & 46.93 & 11.15 & 7.31 & 11.65 & 16.30 & 13.71 \\
                & LAB \cite{wu2018look} & 7.56 & 28.83 & 6.37 & 6.73 & 7.77 & 13.72 & 10.74 \\
                & SAN \cite{dong2018style} \# & 6.32 & 27.91 & 7.01 & 4.87 & 6.31 & 11.28 & 6.60 \\
                & WING \cite{feng2018wing} * & 6.00 & 22.70 & 4.78 & 4.30 & 7.77 & 12.50 & 7.76 \\
                & HRNet18 \cite{sun2019deep} & 4.64 & 23.01 & 3.50 & 4.72 & 2.43 & 8.29 & 6.34 \\
                & STYLE \cite{qian2019aggregation} \# & 4.08 & 18.10 & 4.46 & 2.72 & 4.37 & 7.74 & 4.40 \\
                & AWING \cite{wang2019adaptive} * & \textbf{2.84} & \textbf{13.50} & \textbf{2.23} & \textbf{2.58} & 2.91 & 5.98 & \textbf{3.75} \\
                & \textbf{Ours} & 3.04 & 15.95 & 2.86 & 2.72 & \textcolor{blue}{\textbf{1.45}} & \textcolor{blue}{\textbf{5.29}} & {4.01} \\
        \midrule
            \multirow{8}{*}{{AUC @0.1}}
                & CFSS \cite{zhu2015face} & 0.3659 & 0.0632 & 0.3157 & 0.3854 & 0.3691 & 0.2688 & 0.3037 \\
                & DVLN \cite{wu2017leveraging} & 0.4551 & 0.1474 & 0.3889 & 0.4743 & 0.4494 & 0.3794 & 0.3973 \\
                & HRNet18 \cite{sun2019deep} & 0.5237 & 0.2506 & 0.5102 & 0.5326 & 0.5445 & 0.4585 & 0.4515 \\
                & LAB \cite{wu2018look} & 0.5323 & 0.2345 & 0.4951 & 0.5433 & 0.5394 & 0.4490 & 0.4630 \\
                & SAN \cite{dong2018style} \# & 0.5355 & 0.2355 & 0.4620 & 0.5552 & 0.5222 & 0.4560 & 0.4932 \\
                & WING \cite{feng2018wing} * & 0.5504 & 0.3100 & 0.4959 & 0.5408 & 0.5582 & 0.4885 & 0.4932 \\
                & AWING \cite{wang2019adaptive} * & 0.5719 & 0.3120 & 0.5149 & 0.5777 & 0.5715 & 0.5022 & 0.5120 \\
                & STYLE \cite{qian2019aggregation} \# & \textbf{0.5913} & 0.3109 & 0.5490 & \textbf{0.6089} & 0.5812 & 0.5164 & \textbf{0.5513} \\
                & \textbf{Ours} & 0.5893 & \textbf{0.3150} & \textbf{0.5663} & 0.5953 & \textcolor{blue}{\textbf{0.6038}} & \textcolor{blue}{\textbf{0.5235}} & 0.5329 \\
        \bottomrule
      \end{tabular}
  }
\label{tab:wflw}
\end{table*}

\noindent\textbf{Implementation Details:} Following previous studies, we crop and resize facial images into $256 \times 256$ based on the provided bounding boxes. We follow \cite{chen2019cephalometric} to resize the Cephalometric X-rays to $640 \times 800$. For the Hand and Pelvic X-rays, we resize each image into $512 \times 512$ preserving the original height and width ratio by padding zero values to the empty regions. The proposed model is implemented in PyTorch and is experimented on a single NVIDIA Titan V GPU. We choose $\lambda_1 = \lambda_2 = 1$ for different parts in the overall loss function. HRNet18 \cite{sun2019deep} pretrained on ImageNet is used as our backbone network to extract visual feature maps for its parallel multi-resolution fusion mechanism and deep network design which fits our need for both high resolution and semantic feature representation. The last output after fusion is extracted as feature map of dimension $H\in \!R^{256 \times 64 \times 64}$. We employ 4 residual GCN blocks~\cite{ling2019fast,li2019can} in GCN-global and GCN-local and perform 3 iterations of GCN-local. Adjacency matrix values are initialized to 1/N so that the total weight for each node is 1 to avoid message explosion. 

\begin{table}[t!]
\parbox{.475\textwidth}{
  \centering
    \caption{Evaluation on 300W Common set, Challenge set and Fullset.}
    \resizebox{0.95\linewidth}{!}{
      \begin{tabular}{ccccc}
        \toprule 
            \multicolumn{5}{c}{Inter-Pupil Normalization} \\
        \cmidrule{1-5}
                Method & Year & Comm. & Challenge & Full. \\
        \cmidrule{1-5}
                CFAN \cite{zhang2014coarse} & 2014 & 5.50 & 16.78 & 7.69 \\
                ESR \cite{cao2014face} & 2014 & 5.28 & 17.00 & 7.58 \\
                SDM \cite{xiong2013supervised} & 2013 & 5.57 & 15.40 & 7.52 \\
                3DDFA \cite{zhu2016face} & 2016 & 6.15 & 10.59 & 7.01 \\
                LBF \cite{ren2014face} & 2014 & 4.95 & 11.98 & 6.32 \\
                CFSS \cite{zhu2015face} & 2015 & 4.73 & 9.98 & 5.76 \\
                SeqMT \cite{honari2018improving} & 2018 & 4.84 & 9.93 & 5.74 \\
                TCDCN \cite{zhang2015learning} & 2015 & 4.80 & 8.60 & 5.54 \\
                RCN \cite{honari2016recombinator} & 2016 & 4.67 & 8.44 & 5.41 \\
                TSR \cite{lv2017deep} & 2017 & 4.36 & 7.56 & 4.99 \\
                DVLN \cite{wu2017leveraging} & 2017 & 3.94 & 7.62 & 4.66 \\
                HG-HSLE \cite{zoulearning} & 2019 & 3.94 & 7.24 & 4.59 \\
                DCFE \cite{valle2018deeply} & 2018 & 3.83 & 7.54 & 4.55 \\
                STYLE \cite{qian2019aggregation} \# & 2019 & 3.98 & 7.21 & 4.54 \\
                AWING \cite{wang2019adaptive} * & 2019 & 3.77 & \textbf{6.52} & 4.31 \\
                LAB \cite{wu2018look} & 2018 & 3.42 & 6.98 & 4.12 \\
                WING \cite{feng2018wing} * & 2018 & \textbf{3.27}& 7.18 & \textbf{4.04} \\
                \textbf{Ours} & 2020 & 3.64 & 6.88 & 4.27 \\
        \midrule
            \multicolumn{5}{c}{Inter-Ocular Normalization} \\
        \cmidrule{1-5}
            Method & Year & Comm. & Challenge & Full. \\
        \cmidrule{1-5}
                PCD-CNN \cite{kumar2018disentangling} & 2018 & 3.67 & 7.62 & 4.44 \\
                ODN \cite{zhu2019robust} & 2019 & 3.56 & 6.67 & 4.17 \\
                CPM+SBR \cite{dong2018supervision} & 2018 & 3.28 & 7.58 & 4.10 \\
                SAN \cite{dong2018style} \# & 2018 & 3.34 & 6.60 & 3.98\\
                STYLE \cite{qian2019aggregation} \# & 2019 & 3.21 & 6.49 & 3.86 \\
                LAB \cite{wu2018look} & 2018 & 2.98 & 5.19 & 3.49 \\
                HRNet18 \cite{sun2019deep} & 2019 & 2.91 & 5.11 & 3.34 \\
                HG-HSLE \cite{zoulearning} & 2019 & 2.85 & 5.03 & 3.28 \\
                LUVLi \cite{kumar2020luvli} & 2020 & 2.76 & 5.16 & 3.23 \\
                AWING \cite{wang2019adaptive} * & 2019 & 2.72 & \textbf{4.52} & 3.07 \\
                \textbf{Ours} & 2020 & \textbf{2.62} & 4.77 & \textbf{3.04}\\
        \bottomrule
      \end{tabular}
      \label{tab:300w}
      }
      }
      \quad
      \parbox{.455\linewidth}{
      \centering
        \caption{Evaluation on 300W and COFW-68 testsets with the model trained on 300W training set. 
        }
          \resizebox{0.95\linewidth}{!}{
          \begin{tabular}{cccc}
            \toprule 
                \multicolumn{4}{c}{300W} \\
            \cmidrule{1-4}
                Method & Year & AUC@0.1 & FR@0.1 \\
            \cmidrule{1-4}
                    Deng \textit{et al.} \cite{deng2016m3} & 2016 & 0.4752 & 5.50 \\
                    Fan \textit{et al.} \cite{fan2016approaching} & 2016 & 0.4802 & 14.83 \\
                    DensReg+DSM \cite{alp2017densereg} & 2017 & 0.5219 & 3.67 \\
                    JMFA \cite{deng2019joint} & 2019 & 0.5485 & 1.00 \\
                    LAB \cite{wu2018look} & 2018 & 0.5885 & 0.83 \\
                    HRNet18 \cite{sun2019deep} & 2019 & 0.6041 & 0.66 \\
                    AWING \cite{wang2019adaptive} * & 2019 & \textbf{0.6440} & 0.33 \\
                        \textbf{Ours} & 2020 & 0.6361 & \textbf{0.33} \\
            \midrule
                \multicolumn{4}{c}{COFW-68} \\
            \cmidrule{1-4}
                Method & Year & Mean Error \% & FR@0.1 \\
            \cmidrule{1-4}
                CFSS \cite{zhu2015face} & 2015 & 6.28 & 9.07 \\
                HRNet18 \cite{sun2019deep} & 2019 & 5.06 & 3.35 \\
                LAB \cite{wu2018look} & 2018 & 4.62 & 2.17 \\
                STYLE \cite{qian2019aggregation} \# & 2019 & 4.43 & 2.82 \\
                    \textbf{Ours} & 2020 & \textbf{4.22} & \textbf{0.39} \\
            \bottomrule
          \end{tabular}
          \label{tab:crossvalid}
         }
     
      \centering
        \caption{Evaluations on the hand X-ray and pelvic X-ray images.}
        \resizebox{0.931\linewidth}{!}{
          \begin{tabular}{ccccc}
            \toprule 
                \multicolumn{5}{c}{Hand X-ray Dataset} \\
            \cmidrule{1-5}
               Method & Year & MRE (pix) & Hausdorff & STD\\
            \cmidrule{1-5}
                    HRNet18 \cite{sun2019deep} & 2019 & 12.79 & 26.36 & 6.07 \\
                    Chen \textit{et al.} \cite{chen2019cephalometric} & 2019 & 7.14 & 18.71 & 14.43 \\
                    Payer \textit{et al.}~\cite{Payer2019a} & 2019 & 6.11 & 16.55 & 4.01 \\
                    \textbf{Ours} & 2020 & \textbf{5.57} & \textbf{14.83} & \textbf{3.63} \\
            \midrule
                \multicolumn{5}{c}{Pelvic X-ray Dataset} \\
            \cmidrule{1-5}
               Method & Year & MRE (pix) & Hausdorff & STD\\
            \cmidrule{1-5}
                    HRNet18 \cite{sun2019deep} & 2019 & 24.77 & 71.31 & 19.98 \\
                    Payer \textit{et al.}~\cite{Payer2019a} & 2019 & 20.96 & 68.19 & 21.93 \\
                    Chen \textit{et al.} \cite{chen2019cephalometric} & 2019 & 20.10 & 59.92 & 20.14 \\
                    \textbf{Ours} & 2020 & \textbf{18.39} & \textbf{56.72} & \textbf{17.67} \\
            \bottomrule
          \end{tabular}
    }
  \label{tab:hand_pelvic}
}
\end{table}

\subsection{Comparison with the SOTA methods}
\noindent\textbf{WFLW:} WFLW is a comprehensive public facial landmark detection dataset focusing on multi-discipline and difficult detection scenarios. Summary of results is shown in Table~\ref{tab:wflw}. Following previous works, three evaluation metrics are computed: Mean Error, FR@0.1 and AUC@0.1. Our model achieves \textit{4.21\%} mean error which outperforms all the strong state-of-the-art methods including AWING \cite{wang2019adaptive} which adopts a new adaptive loss function, SAN \cite{dong2018style} and STYLE \cite{qian2019aggregation} which leverage additional generated images for training. The most significant improvements lie in \textit{Make-up} and \textit{Occlusion} subsets, where only partial landmarks are visible. Our model is able to accurately infer those hard cases based on the visible landmarks due to the benefit of preserving and leveraging graph structural knowledge. This can be further illustrated by examining the visualization results for the occlusion scenarios in Figure~\ref{fig:face_visual}.

\noindent\textbf{300W:} There are two evaluation protocols, namely inter-pupil and inter-ocular normalizations. In this paper, we conduct experiments under both settings on the detection results in order to comprehensively evaluate with the other state-of-the-arts. As can be seen from Table~\ref{tab:300w}, our model achieves competitive results in both evaluation settings comparing to the previous best models, STYLE \cite{qian2019aggregation}, LAB \cite{wu2018look} and AWING \cite{wang2019adaptive} which are all heatmap-based. Comparing to the latest coordinate-based model ODN \cite{zhu2019robust} and DVLN \cite{wu2017leveraging}, our method achieves improvements in large margins (\textit{27\%} and \textit{8\%} respectively) which sets a remarkable milestone for coordinate-based models, closing the gap between coordinate- and heatmap-based methods.

\noindent\textbf{COFW-68 and 300W testset:} To verify the robustness and generalizability of our model, we conduct inference on images from COFW-68 and 300W testset using the model trained on 300W training set and validated on 300W fullset. Results summarized in Table~\ref{tab:crossvalid} indicating our model's superior performance over most of the other state-of-the-art methods in both datasets. In particular for the COFW-68 dataset, the Mean Error and FR@0.1 are significantly improved (\textit{5\%} and \textit{86\%}) comparing to the previous best model, STYLE \cite{qian2019aggregation}, demonstrating a strong cross-dataset generalizability of our method.

\begin{table*}[t]

\centering
\centering
\resizebox{0.8\textwidth}{!}{\begin{minipage}{\textwidth}
\centering
    \caption{Evaluation on the public Cephalometric dataset.}
      \begin{tabular}{cccccccccccc}
        \toprule 
            \multirow{2}{*}{Model} & \multirow{2}{*}{Year}& \multicolumn{5}{c}{Validation set} & \multicolumn{5}{c}{Test set} \\
            \cmidrule(lr){3-7} \cmidrule(lr){8-12}
            & & MRE & 2mm & 2.5mm & 3mm & 4mm & MRE & 2mm & 2.5mm & 3mm & 4mm \\
            \cmidrule{1-12}
            Arik \textit{et al.} \cite{arik2017fully} & 2017 & - & 75.37 & 80.91 & 84.32 & 88.25 & - & 67.68 & 74.16 & 79.11 & 84.63 \\
            HRNet18 \cite{sun2019deep} & 2019 & 1.59 & 78.11 & 86.81 & 90.88 & 96.74 & 1.84 & 69.89 & 78.95 & 85.16 & 92.32 \\
            Payer \textit{et al.} \cite{Payer2019a} & 2019 & 1.34 & 81.47 & 89.36 & 93.15 & 97.01  & 1.65 & 69.94 & 78.84 & 85.74 & 93.89 \\
            Chen \textit{et al.} \cite{chen2019cephalometric} & 2019 & 1.17 & 86.67 & 92.67 & 95.54 & \textbf{98.53} & 1.48 & 75.05 & 82.84 & \textbf{88.53} & \textbf{95.05} \\
            \textbf{Ours}  & - & \textbf{1.04} & \textbf{88.49} & \textbf{93.12} & \textbf{95.72} & 98.42 & \textbf{1.43} & \textbf{76.57} & \textbf{83.68} & 88.21 & 94.31 \\
        \bottomrule
      \end{tabular}
      \label{tab:cepha}
  \end{minipage}}
\end{table*}

\noindent\textbf{Cephalometric X-rays:} We further applied our model on a public Cephalometric X-ray dataset and compare with HRNet18 \cite{sun2019deep} and three domain specific state-of-the-art models on this dataset, Arik \textit{et al.} \cite{arik2017fully}, Payer {et al.} \cite{Payer2019a} and Chen \textit{et al.}~\cite{chen2019cephalometric}. As is shown in Table~\ref{tab:cepha}, our model significantly outperforms Arik \textit{et al.}, HRNet18~\cite{sun2019deep} and Payer {et al.} \cite{Payer2019a} in all metrics. Comparing to Chen \textit{et al.}~\cite{chen2019cephalometric}, we also achieve improved overall accuracy evaluated under MRE. A closer look at the error distribution reveals that our model is able to achieve more precise localization under smaller error ranges, i.e., 2mm and 2.5mm. 

\noindent\textbf{Hand and Pelvic X-rays:} As shown in Table~\ref{tab:hand_pelvic}, our model achieves susbstantial performance improvements comparing to the HRNet18 \cite{sun2019deep}, Payer {et al.} \cite{Payer2019a} and Chen \textit{et al.} \cite{chen2019cephalometric} on both the Hand and Pelvic X-ray datasets. On Hand X-ray, where the bone structure can vary in different shapes depending on the hand pose, our method still achieves largely reduced Hausdorff distance as well as its standard deviation, reveling DAG's ability in capturing landmark relationships under various situations toward robust landmark detection.

\noindent\subsection{Graph Structure Visualization}
\begin{figure*}[t]
	\centering
	\includegraphics[width=\textwidth]{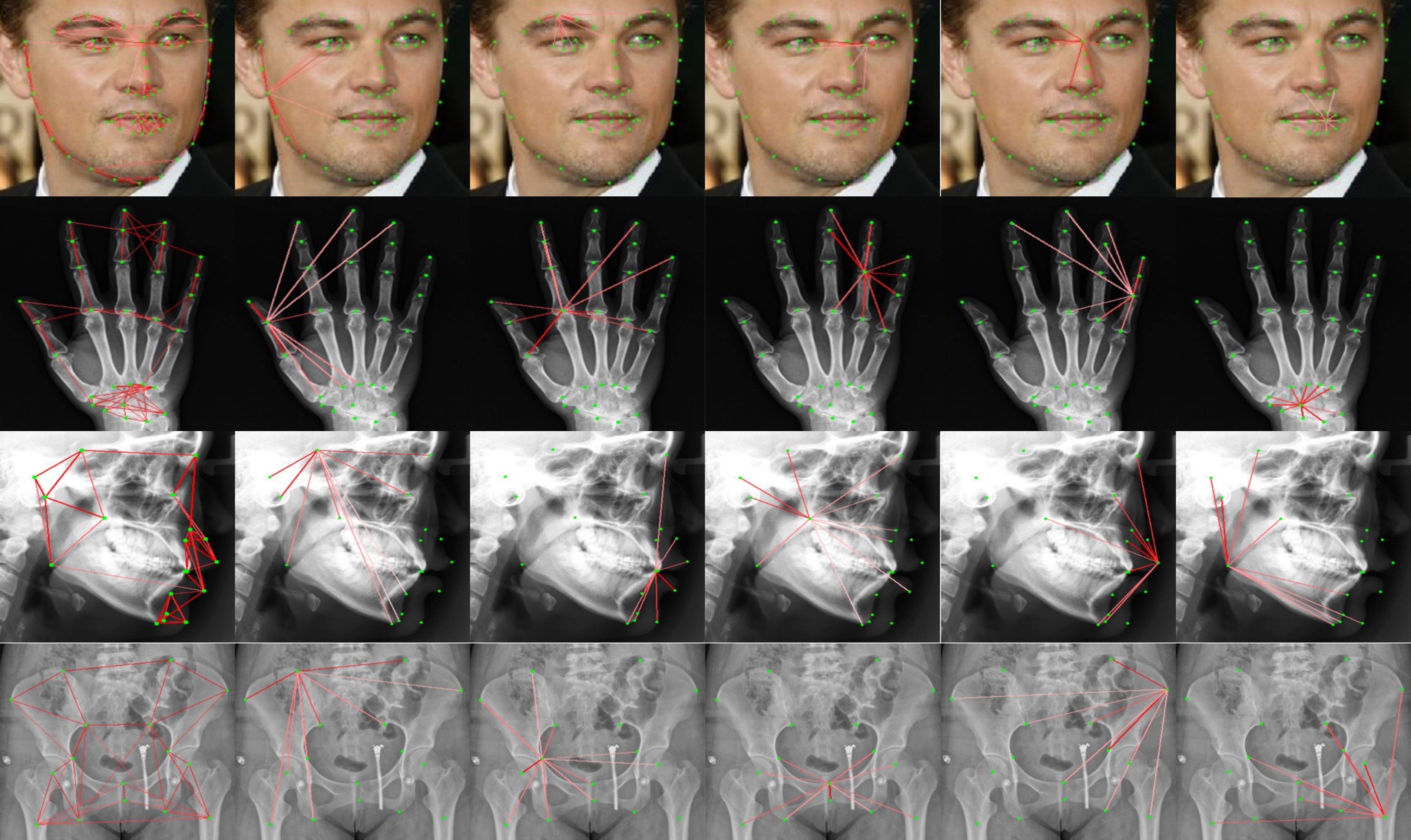}
	\caption{Graph structure visualization. Red lines: edges. Green dots: landmarks. Deeper red means higher edge weights. \textbf{[Leftmost column]}: the constructed graphs (3 highest weighted edges for each landmark). \textbf{[Right 5 columns}]: for the 5 landmarks, the most related neighbors (10 highest weighted edges).}
	\label{fig:graph_vis}
\end{figure*}

To better understand learning outcomes, we look into the visualization on the learned graph structure. As shown in Figure~\ref{fig:graph_vis}, the learned structures in different domains are meaningful indicating strong connections between 1) spatially close landmarks, and 2) remote but related landmarks that move coherently, e.g. symmetrical body parts. We believe the mechanism behind our algorithm is relying on these locations to provide reliable inductions when it makes movement predictions, such as similar movements by neighbors, or fixed spatial relationships by the symmetrical body parts (e.g., eyes, pelvis). With the learnable graph connectivity, we are able to capture the underlying landmarks relationships for different objects.

\subsection{Ablation Studies} 
In this section, we examine the performance of the proposed methods by conducting ablation studies on the 300W fullset. We analyze: 1) the overall effect of using the proposed DAG to regress landmark coordinates, 2) the individual effect of learning the graph connectivity, 3) the individual effect of incorporating shape feature into the graph signal. More ablation studies can be found in the supplementary material.

\begin{table}[htbp]
  \centering
  \resizebox{0.9\textwidth}{!}{\begin{minipage}{\textwidth}
    \centering
    \captionof{table}{Ablation studies on the effectiveness of the proposed method DAG.}
    \begin{tabular}{lcccc}
    \toprule 
                         & VGG16 & ResNet50 & StackedHG4 & HRNet18 \\ \hline
        Global feature   & 4.66    & 4.33      & 4.31 &  4.30 \\
        Local feature    & 4.42    & 4.10      & 3.96 &  3.72 \\
        Proposed DAG     & \textbf{3.66} & \textbf{3.65} & \textbf{3.07} & \textbf{3.04} \\
    \bottomrule
    \end{tabular}
    \label{tab:ablation_backbone}
    
    \setlength{\tabcolsep}{10pt}
    \captionof{table}{Ablation study on graph connectivity and shape feature.}
    \begin{tabular}{ccc}
    \toprule
              &   w.o Shape Feature    & w. Shape Feature \\
    \midrule
    Self     &  3.31      &  3.16    \\
    Uniform  &  3.16     &  3.12    \\
    Learned & \textbf{3.08}     &  \textbf{3.04} \\
    \bottomrule
    \end{tabular}
    \label{tab:connect}
  \end{minipage}}
\end{table}

\textbf{Overall effect of the proposed DAG:} We analyze the effect of using DAG to regress landmark coordinates in comparison with two baselines, namely 1) \textit{Global feature}: The last feature map of the backbone network is global average pooled to produce a feature vector, which connects to a fully connected layer to regress landmark coordinates. This approach is similar to previous coordinate regression based methods, e.g.~\cite{wu2017leveraging,zhang2015learning}. 2) \textit{Local feature}: The feature vectors are interpolated at each landmark's initial location on the last feature map of the backbone CNN. Then each landmark's feature vector is connected to a fully connected layer to regress the landmark's coordinate. To decouple the effect of the backbone strength, each experiment is conducted on four popular landmark detection backbone networks, namely VGG16~\cite{simonyan2014very}, ResNet50~\cite{he2016deep}, StackedHourGlass4~\cite{newell2016stacked}, HRNet18~\cite{sun2019deep}. Results are listed in Table~\ref{tab:ablation_backbone}. By comparing different regression methods with the same backbone (columnwise), DAG achieves the best results indicating the proposed framework's strong localization ability. By comparing DAG's results under different backbones (last row), we observe DAG's consistent performance boost demonstrating its effectiveness and promising generalizability.

\textbf{Individual effect of learning graph connectivity:}
We study three kinds of graph connectivity schemes, namely 1) \textit{Self}-connectivity: The landmarks only connect to themselves and no other landmarks. 2) \textit{Uniform} connectivity: The landmarks connects to all other landmarks using the same edge weight. 3) \textit{Learned} connectivity: learned edge weights as proposed. As summarized in Table~\ref{tab:connect}, regardless of using shape feature or not, using uniform connectivity performs results in better performance than self-connectivity, demonstrating the importance of allowing information exchange on the graph. The learned connectivity performance the best, further demonstrating that learned edge weights further improve the effectiveness of information exchange on the graph. 
\textbf{Individual effect of incorporating shape feature:}
We analyze the effect of incorporating the shape feature using self, uniformed and learned connectivities, respectively. As shown in Table~\ref{tab:connect}, on all three types of connectivities, incorporating the proposed shape feature into graph signal results in improved performance especially for self-connective graphs, where the shape feature adds the missing global structure information.

\section{Conclusion} In this paper, we introduce a robust and accurate landmark detection model named Deep Adaptive Graph (DAG). The proposed model deploys an initial landmark graph, and then deforms and progressively updates the graph by learning the adjacency matrix. Graph convolution operations follow the strong structural prior to enable effective local information exchange as well as global structural constraints for each step's movements. The superior performances on three public facial image datasets and three X-ray datasets prove both the effectiveness and generalizability of the proposed method in multiple domains.

\subsubsection{Acknowledgement.} This work is supported in part by NSF through award IIS-1722847, NIH through the Morris K. Udall Center of Excellence in Parkinson's Disease Research. The main work was done when Weijian Li was a research intern at PAII Inc.

\newpage
\section{Additional discussions with related works:} 
Though some recent works~\cite{tompson2014joint,cao2017realtime,Payer2019a} propose to model landmark relationship, our problem/method has large differences from them. Tompson \textit{et al.}~\cite{tompson2014joint} propose to use spatial information in a post-processing step to filter outliers, while we leverage visual-spatial joint features for landmark regression. Also, the PAF proposed by Cao \textit{et al.}~\cite{cao2017realtime} focuses on a different task of assembling detected key points for multi-person parsing. Zhao \textit{et al.}~\cite{zhao2019semantic} focus differently on predicting 3D poses from 2D joints. Their 2D joints are generated by a pre-trained 2D pose estimation network. Besides, their network structure is predefined by a fixed adjacency matrix while we actively learn the structures. Payer \textit{et al.}~\cite{Payer2019a}, propose a spatial configuration branch to disambiguate candidates from the heatmap predictions. There is no explicit landmark structure modeling. In contrast, we explicitly model shape through a graph representation with learnable connectivity.

Among the SOTA, WING~\cite{feng2018wing} is pure coordinate-based, while LAB~\cite{wu2018look} and AWING~\cite{wang2019adaptive} integrate face boundary information via heatmap, which is their key contributions. The gap between WING and AWING is significant on WFLW, which is a more challenging dataset than 300W in terms of dataset scale, pose variations, occlusions, etc. Our method performs significantly better than WING on WFLW by reducing the failure rate by 50\%, and is competitive to AWING. In addition, WING focuses on loss design, which is orthogonal and complementary to our novelty. By employing WING loss in our method, our performance can be further improved (e.g., on 300W, inter-pupil NME from 4.27 to 4.21 and inter-ocular NME from 3.04 to 3.01). While LAB and AWING utilize global representation, human knowledge on face structure via a boundary heatmap is injected, leading to task-specific solutions. In contrast, our method is a general landmark detection method to model the structural information via a self-learned graph structure. 

\begin{figure}[t]
	\centering
	\resizebox{0.8\columnwidth}{!}{\includegraphics{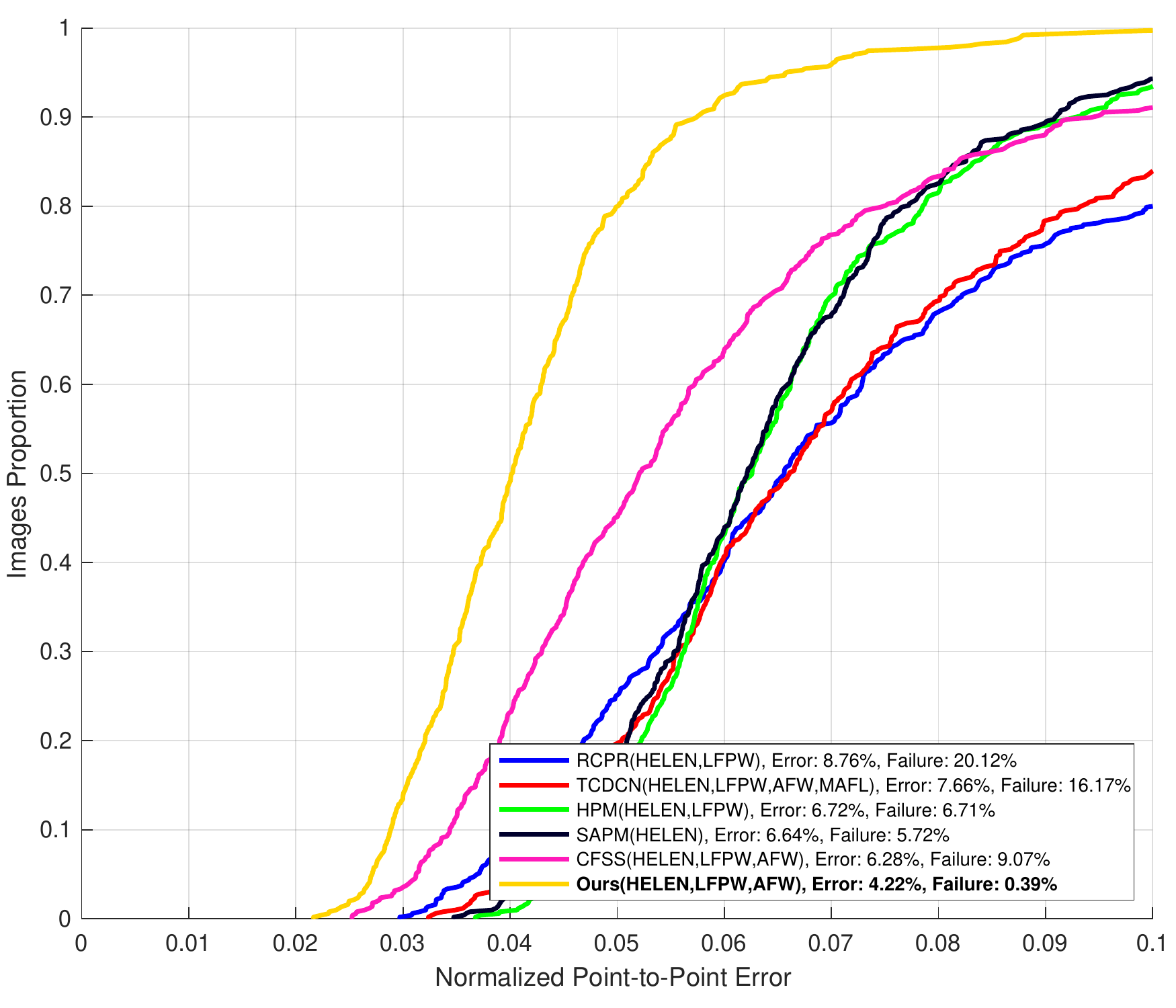}}
	\caption{Cumulative Errors Distribution (CED) curve results on the COFW-68 test set.}
	\label{fig:ced}
\end{figure}

\section{CED Curve:}
Following previous works \cite{wu2018look,qian2019aggregation}, we report Cumulative Errors Distribution (CED) curve result on cross-evaluations of COFW-68 test set. Recall that the success rate measures the proportion of images that have a localization error below a certain threshold \cite{ghiasi2015occlusion}. Thus, given a range of thresholds, the corresponding success rates will form a distribution which is considered as Cumulative Error Distribution (CED). For clearer comparison, we include both Normalized Mean Error (\textbf{Error}) as well as the Failure Rate (i.e. $1-Success Rate$) (\textbf{Failure})  at threshold of 0.1. As we can see from Figure~\ref{fig:ced}, our model outperforms previous methods by a large margin, especially in Failure Rate which is reduced to 0.39\% for the first time. The comparison of numerical NME and Failure Rate values with the other state-of-the-arts can be found in Table 3 in our submitted ECCV-20 main paper.

\section{Ablation Studies} 
Here we conduct three more types of ablation studies, namely: (1) The comparison of the transformation method used in GCN-global. (2) The effectiveness of the proposed GCN modules.(3) The comparison of different number of regression steps used in GCN-local. Results are recorded in Table~\ref{tab:ablations}.
\begin{table}[t]
    \centering
      \captionof{table}{Ablation studies on the proposed model with 300W fullset under Inter-Ocular normalization.}
        \resizebox{1\linewidth}{!}{
            \begin{tabular}{c|c|c|c|c}
            \toprule 
                    Different Transformations & \multicolumn{2}{c|}{Affine Transformation}& \multicolumn{2}{|c} {Perspective Transformation (Ours)}\\
                    NME & \multicolumn{2}{c|}{3.13} & \multicolumn{2}{|c}{\textbf{3.04}} \\
            \midrule 
                    Effectivenes of GCN modules & \multicolumn{2}{c|}{Replace GCN-global with CNN}& \multicolumn{2}{|c} {Replace GCN-local with MLP}\\
                    NME & \multicolumn{2}{c|}{3.12} & \multicolumn{2}{|c}{3.18} \\
            \midrule 
                    Different GCN Steps & Step=1 & Step=3 (Ours) & Step=5 & Step=7 \\
                    NME & 3.24 & \textbf{3.04} & 3.07 & 3.11 \\
            \bottomrule
          \end{tabular}
        \label{tab:ablations}
      }
\end{table}

\textbf{Choice of transformations:} We experiment two types of GCN-global choices: (1) Adopt Affine Transformation. In this case, the performance of our GCN-global module drops to 3.13.(2) Adopt Perspective Transformation. We achieve the best result as 3.04 which is also reported in our main paper. This indicates that GCN-global can better locates ROIs with the more flexible perspective transformation.

\textbf{Effectiveness of GCN modules:} We examine the effectiveness of the proposed GCN modules by: (1) Replacing GCN-global with a CNN block: we replace the GCN-global module with a 2-layer CNN (Conv/BN/ReLU) with Global Average Pooling predicting 9 transformation parameters. The average error increased from 3.04 to 3.12. (2) Replacing GCN-local with a MLP block: we remove the connectivity used in GCN-local, making it a simple MLP (FC/ReLU). The average error increased from 3.04 to 3.18. These indicating the importance of the proposed GCN modules.

\textbf{Number of steps:} We analyze different choices of steps for GCN-local. Results are shown in Table~\ref{tab:ablations}. The overall performance improves as the number of steps increases indicating the benefit of cascading multiple regressions. The best performance is achieved when GCN-local is implemented with three iterations. 

\section{More Settings:}
We describe more settings for training the model. Adam optimizer is adopted with initial learning rate $lr=0.0001$. The learning rate decreases at every 100 epochs. $L2$ penalty is applied to the training parameters with rate $0.0001$. Margin for training GCN-global is set to $m=0.1$ for Face300W, $m=0.15$ for WFLW, $m=0.15$ for three Medical datasets. All data augmentations we used: (1) Rotate input image with a random angle in [-30, 30]. (2) Random flip the input image horizontally. (3) Scale input image with a random factor in [0.75, 1.25].

\clearpage
%
%
\bibliographystyle{splncs04}
\bibliography{mybib}
\end{document}